# Automated Flight Test and System Identification for Rotary Wing Small Aerial Platform using Frequency Responses Analysis

**Widyawardana Adiprawita\***, **Adang Suwandi Ahmad** = **and Jaka Semibiring**+

*School of Electric Engineering and Informatics  
Institut Teknologi Bandung, Bandung, Id.  
e-mail: wadiprawita@stei.itb.ac.id

= School of Electric Engineering and Informatics  
Institut Teknologi Bandung, Bandung, Id.  
e-mail: asa@isrg.itb.ac.id

+ School of Electric Engineering and Informatics  
Institut Teknologi Bandung, Bandung, Id.  
e-mail: jaka@ itb.ac.id

**Abstract**

This paper proposes an autopilot system that can be used to control the small scale rotorcraft during the flight test for linear-frequency-domain system identification. The input frequency swept is generated automatically as part of the autopilot control command. Therefore the bandwidth coverage and consistency of the frequency swept is guaranteed to produce high quality data for system identification. Beside that we can set the safety parameter during the flight test (maximum roll / pitch value, minimum altitude, etc) so the safety of the whole flight test is guaranteed. This autopilot for automated flight test will be tested using hardware in the loop simulator for hover flight condition.

## 1 Introduction

With the ability to take off and land vertically and hover, along with natural agility and controllability, rotorcraft will extend the potential roles for UAVs. Helicopters already have an irreplaceable role among aircraft, indispensable for a variety of tasks ranging from medical evacuation to transportation to construction in confined areas. Such rotorcraft UAVs (RUAV) are already greatly valued by the military for applications for a range of battlefield tasks, such as exploration and even combat operation. There are also numerous examples for civilian applications including filmmaking (allowing both steady and dynamic aerial views), close up inspection (bridges, buildings, dams) and digital terrain modeling (where a small vehicle, because of its potential for closer proximity to the terrain and structures, could gather more detailed features).

Small-size rotorcraft tend to be naturally mode maneuverable and mode responsive than traditional full-scale rotorcraft. The maneuverability of such vehicles can offer a tremendous operational advantages if harnessed during autonomous flight. Taking full advantage of the small-scale helicopter's natural abilities, however, has been a major challenge to its use as a UAV platform.

Simply speaking, helicopters are difficult to fly. They are unstable, requiring continuous attention from the pilot. Four primary inputs are necessary to control their motions: the longitudinal and lateral cyclic for horizontal motions in their respective directions; collective for vertical motion; and pedal for yaw (heading) motion.

### 1.1 Vehicle Dynamics Model

A number of available control design methods could, in theory, be applied to address the rotorcraft control issues (described later in this chapter). To be successfully applied, these methods require accurate vehicle dynamics models.

The physical approach to modeling involves deriving the equations of motion from the ground up, using the fundamental laws of mechanics and aerodynamics. This approach is referred to as first-principle modeling. For a system involving numerous physical effects, the resulting equations of motion are typically high-order nonlinear coupled vector differential equations. A first principles approach requires considerable knowledge of and experience with all the phenomena involved in rotorcraft flight. This makes the first-principles approach difficult to do.

For control design applications, as well as for flying qualities studies, simpler linear models are often sufficient. Linear model have been sued extensively and successfully for rotorcraft. At a precise operating point, and even within a certain region around that point, linear models accurately capture the essential effects of the vehicle dynamics. This is advantageous, since numerous analysis and design techniques are available for linear system. However, to cover the entire flight envelope, multiple linear models and therefore multiple controllers are necessary. An effective approach to deriving accurate linear models of a plant is linear system identification.

**1.2 System Identification and CIFER**

System identification uses experimental input-output data collected from a plant to produce a mathematical representation of the system's dynamics. This method is attractive because it is direct, it is based on real data, and integrates validation and model refinement into the modeling process.

In the 1980s, rotorcraft experts undertook important efforts in developing effective identification methods and tools. CIFER (Comprehensive Identification from FrEquency Responses from the US Army/NASA Rotorcraft Division, is one of today's standard tools for rotorcraft identification).

Advanced high-bandwidth flight control design requires models that are accurate at higher frequencies. A very effective and accurate way to obtain linear rotorcraft models is trough linear-frequency-domain system identification. In this methods, frequency responses are estimated from the collected input-output data. Then, the parameters of a linear model capturing the key vehicle dynamics are identified by tuning them to fit the estimated frequency responses.

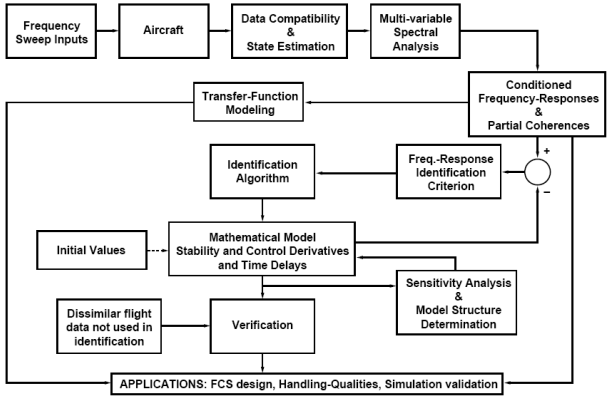

**Figure 1:** Frequency-domain system identification procedure

However, the flight test which is suitable for linear-frequency-domain system identification is difficult to be conducted by human pilot. This especially the case with the small scale rotorcraft. The pilot controls the small scale rotorcraft using visual feedback from safe distance. The delay of response based from visual feedback make the pilot difficult to produce the needed quality of frequency input-output data. Beside that human pilot can not accurately covers the whole frequency needed for the test.

This paper proposes an autopilot system that can be used to control the small scale rotorcraft during the flight test for linear-frequency-domain system identification.

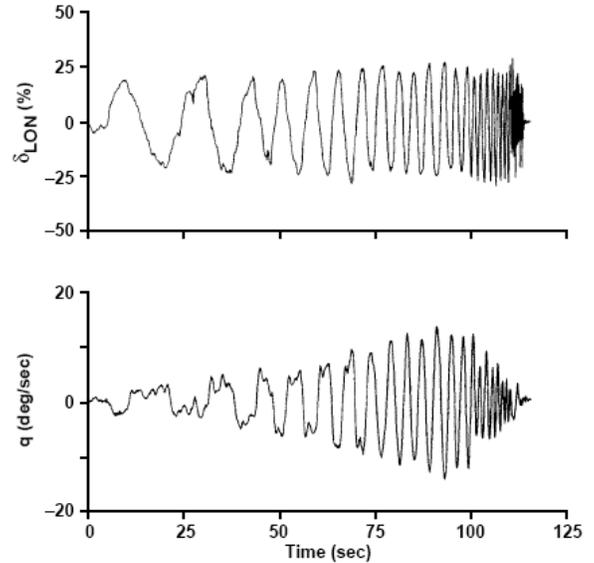

**Figure 2:** Typical input output pair needed for frequency domain system identification

## 2  Methodology

The whole experimentation in this research is conducted in X-Plane simulation. This approach is taken based on previous successful development of fixed wing autopilot based on X-Plane hardware in the loop simulation. So the big picture of the methodology are :

1. Development of PID based rotary wing autopilot
2. Development of frequency-sweep input algorithm
3. Development of simulation system in X-Plane
4. Processing the simulation data in CIFER
5. Hardware design of automated flight test system
6. Hardware in the loop (HIL) simulation
7. Processing the HIL simulation data in CIFER
8. Real world experimentation to get flight data
9. Processing the real world experimentation data in CIFER
10. Optimal Multiple Input Multiple Output (MIMO) control system development based on vehicle's model identified in CIFER
11. Real world experimentation to test Optimal MIMO control system

The research has been completed until the 4th step, so step 1 to 4 will be presented in this paper. The remaining steps hopefully will be completed in near future.

## 3  PID Based Rotary Wing Autopilot

A safe controller for rotary wing platform is needed to control the platform during frequency swept flight test. It has been mentioned before that frequency-domain system identification is difficult to be conducted by human pilot. For now, only hover condition will be implemented in the autopilot. Forward flight condition will be implemented in future development.

It has been shown in several previous research conducted by other institution that PID based controller can be used to control small rotary wing platform. This controller is developed based on empirical approach without any mathematical model. Of course the controller is not optimal (because no mathematical can be used to justify this optimality), but this kind of controller proved to be relatively safe.

The objective of this research is to developed a rotary wing mathematical model based on flight test, and not first principle approach. So PID based controller is the logical choice for the automated flight test. Based on empirical experience, a PID cascaded controller is implemented.

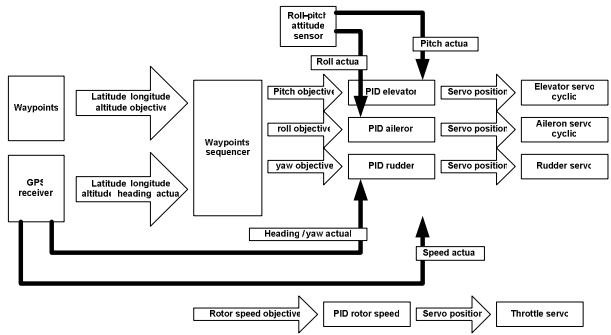

**Figure 3:**  PID cascaded controller for rotary wing

This PID controller is tested in X-Plane simulation environment. There are two main reason why X-Plane is used :

1. X-Plane is very interesting for non aerodynamicist developer, because we can make an airframe based only on its geometric dimension. The physics model is based on a process called Blade Element Theory. This set of principles breaks an airframe down by geometric shape and determines the number of stress points along its hull and airfoils. Factors such as drag coefficients are then calculated at each one of these areas to ensure the entire plane is being affected in some way by external forces. This system produces figures that are far more accurate than those achieved by taking averages of an entire airfoil, for example. It also results in extremely precise physical properties that can be computed very quickly during flight, ultimately resulting in a much more realistic flight model. The X-Plane accuracy of the flight model is already approved by FAA, for full motion simulator to train commercial airline pilot.

2. X-Plane's functionality can be customized using a plug in. A plug in is executable code that runs inside X-Plane, extending what X-Plane does. Plug ins are modular, allowing developers to extend the simulator without having to have the source code to the simulator. Plug ins allow the extension of the flight simulator's capabilities or gain access to the simulator's data.

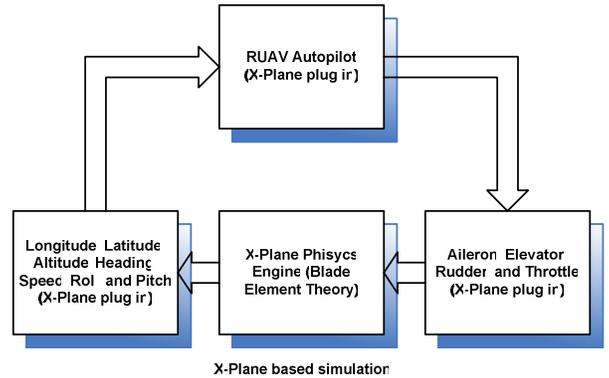

**Figure 4:**  X-Plane based simulation

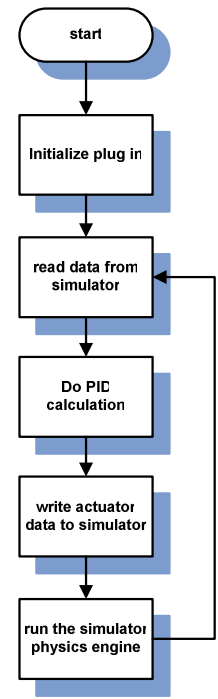

**Figure 5:**  PID controller flow chart

After spending several hours of tuning the gain parameters, the cascaded PID controller can hover the small rotorcraft (RC-02) in X-Plane reliably.

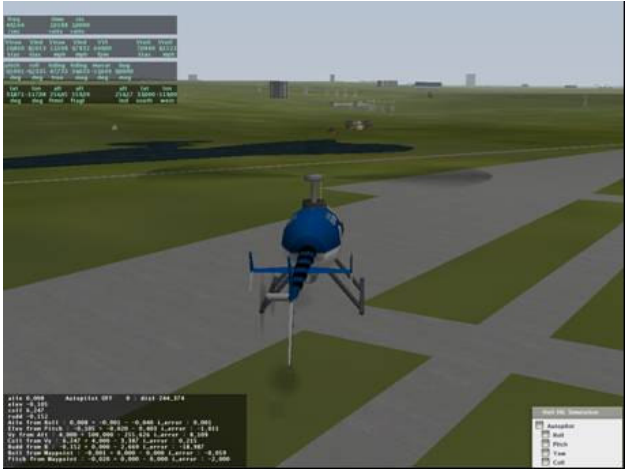

**Figure 6:** PID controller hovering RC-02 in X-Plane

### 4 Automated frequency-sweep input

Reference [] mentioned that computer-generated frequency-sweep input can be both effective and time efficient in many flight application, but they must be properly synthesized in order to obtain the desired frequency spectrum. Tischler have developed an automated frequency-sweep testing approach, using an exponential sweep and white noise. Here is the formulation of the automated frequency-sweep input :

$$\omega = \omega_{min} + K(\omega_{max} - \omega_{min}) \qquad (1)$$

where

$$K = C_2[\exp(\frac{C_1 t}{T_{rec}}) - 1] \qquad (2)$$

The previously developed PID controller need to be modified to accommodate the this frequency-sweep input. Here is the modification needed :

1. Frequency-sweep input need to conducted in four input channel (roll/lateral cyclic, pitch/longitudinal cyclic, yaw or pedal and collective axis)

2. PID controller need to be active on all control axis except the one where the frequency-sweep input is conducted

3. There will be several safety criteria (maximum absolute roll angle, maximum absolute pitch angle, minimum and maximum altitude, and maximum absolute yaw rate) during frequency-sweep input. If one of the safety criteria is violated, the frequency-sweep input is aborted and the PID controller will be activated on all control axis to bring the platform back into steady hover condition.

4. All flight parameter need to be logged into CIFER text input format.

### 5 Data Processing using CIFER

Frequency responses fully describe the linear dynamics of a system. When the system has nonlinear dynamics (to some extent all real physical systems do), system identification determines the describing functions which are the best linear fit of the system response based on a first harmonic approximation of the complete Fourier series. For the identification, we used a frequency domain method, developed by the U.S. Army and NASA, known as CIFER (Comprehensive Identification from Frequency Responses). While CIFER was developed specifically for rotorcraft applications, it has been successfully used in a wide range of fixed wing and rotary wing, and unconventional aircraft applications. CIFER provides a set of utilities to support the various steps of the identification process. All the tools are integrated around a database system which conveniently organizes the large quantity of data generated throughout the identification process. In this research we are using CIFER Student Edition.

The steps involved in the identification process are:

1. Collection of flight data. The flight-data is collected during special flight experiments using frequency sweeps. This step is completed using data logged from X-Plane simulation.

2. Frequency response calculation. The frequency response for each input-output pair is computed using a Chirp-Z transform. At the same time, the coherence function for each frequency response is calculated. This step is completed using FRESPID utility from CIFER.

3. Multivariable frequency domain analysis. The single-input single-output frequency responses are conditioned by removing the effects from the secondary inputs. The partial coherence measures are computed. This step is completed using MISOSA utility from CIFER.

4. Window Combination. The accuracy of the low and high frequency ends of the frequency responses is improved through optimal combination of frequency responses generated using different window lengths. This step is completed using COMPOSITE utility from CIFER.

5. State-space identification. The parameters (derivatives) of an a priori-defined state-space model are identified by solving an optimization problem driven by frequency response matching. This step is completed using DERIVID utility from CIFER. Due to limitation in the CIFER Student Edition (5 state, 2 input), we have to separate lateral and longitudinal dynamics.

6. Time Domain Verification. The final verification of the model accuracy is performed by comparing the time responses predicted by the model with the actual helicopter responses collected from flight experiments using doublet control inputs. This step is completed using VERIFY utility from CIFER.

**5.1 Frequency response calculation**

The flight test simulation in X-Plane produced four kind of frequency-sweep data for each of control axis (lateral, longitudinal, rudder/pedal and collective). Data is recorded in 50Hz frequency. After flight test the data is processed using FRESPID. Here we only present one example from those 4 control axis (the other are typically same).

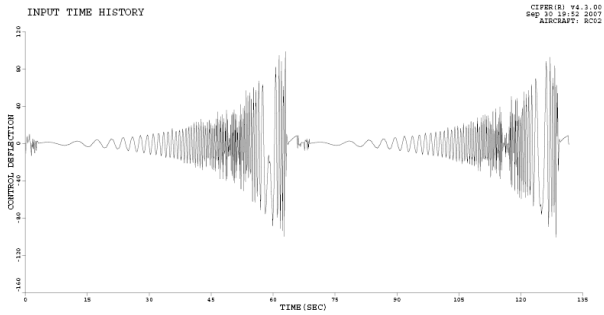

**Figure 7:** Frequency-sweep input (lateral axis)

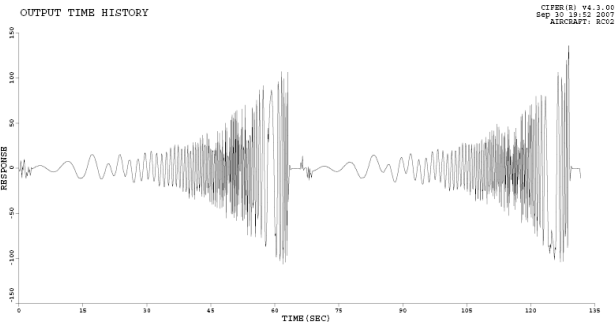

**Figure 8:** Frequency-sweep output P / roll rate (lateral axis)

To measure the frequency coverage of the frequency sweep-input we can see the input auto spectrum plots. This is the same as probability density function. The following figure can give us the illustration of how good the coverage of the automated frequency-sweep input algorithm.

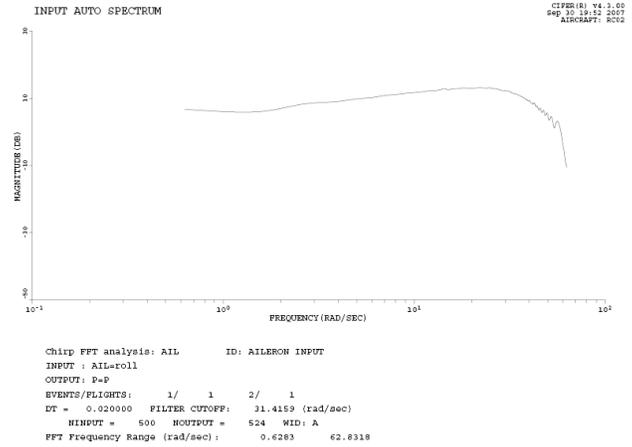

**Figure 9:** Frequency-sweep input auto spectrum (lateral axis)

The frequency response (magnitude and phase response) for each input-output pair is computed using a Chirp-Z transform in this step. This can be viewed as single-input single-output characteristic of the rotary wing platform.

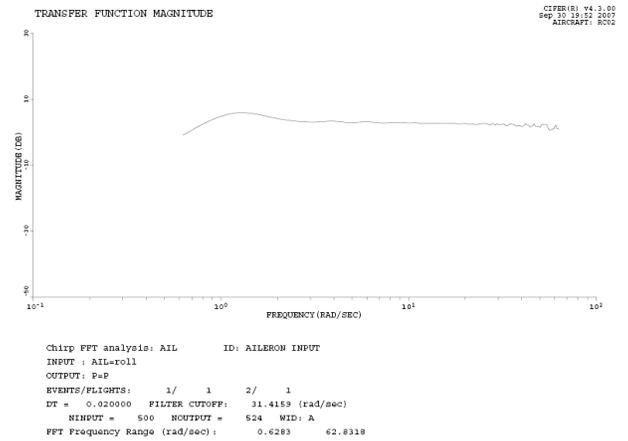

**Figure 10:** Magnitude frequency response (lateral axis)

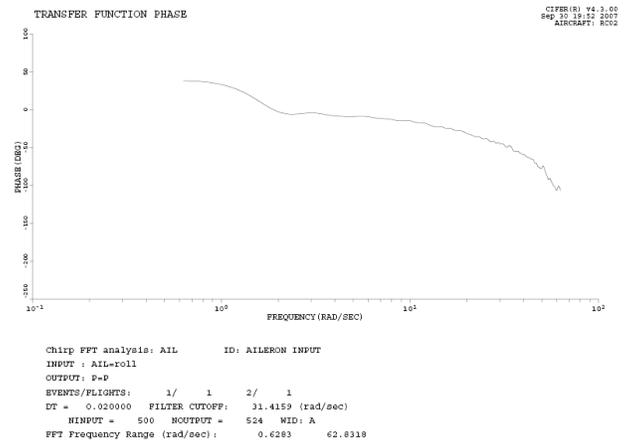

**Figure 11:** Phase frequency response (lateral axis)

A key metric to verify that the flight data is satisfactory for the purpose of system identification is its coherence. Typically coherence value of larger than 0.6 is expected.

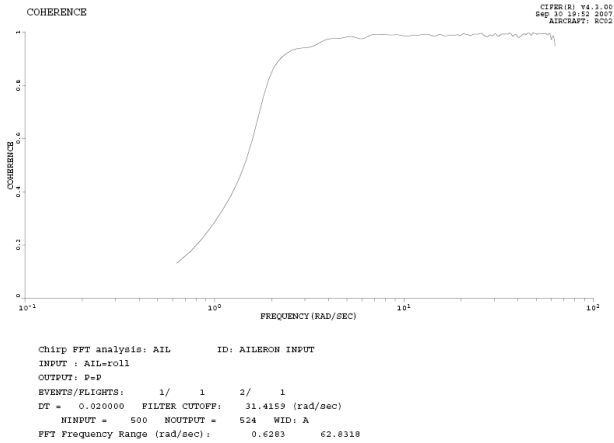

**Figure 12:** Coherence value (lateral axis)

### 6 State-space identification

After conducting single-input single-output frequency responses conditioning by removing the effects from the secondary inputs using MISOSA and optimally combining frequency responses generated with different window lengths using COMPOSITE, we can start constructing the state space model.

The state space form used by CIFER are :

$$M \dot{x} = Fx + Gu(t-\tau) \quad (3)$$

And the measurement output

$$y = Hx + Ju(t-\tau) \quad (4)$$

In this research only lateral response will be studied and only basic forces will be used as state variable. M is identity matrix. J is zero matrix. H is identity matrix. and

$$x = \begin{bmatrix} P \\ R \\ Ay \end{bmatrix} \quad (5)$$

Where $\tau$ is zero, so

$$u = \begin{bmatrix} aileron \\ rudder \end{bmatrix} \quad (6)$$

So we only need to find F and G matrix. We use DERIVID utility to find F and G matrix. Here is the result :

$$F = \begin{bmatrix} -.6411E+02 & 0.0000E+00 & 0.3766E+02 \\ 0.0000E+00 & -.6803E+02 & 0.0000E+00 \\ 0.6056E+00 & 0.0000E+00 & -.5749E+00 \end{bmatrix} \quad (7)$$

$$G = \begin{bmatrix} 0.8700E+02 & 0.1000E+01 \\ 0.1000E+01 & 0.1713E+03 \\ -.4814E+00 & 0.1000E+01 \end{bmatrix} \quad (8)$$

Then we can verify the resulting state space model using flight data and VERIFY utility. Here are the result :

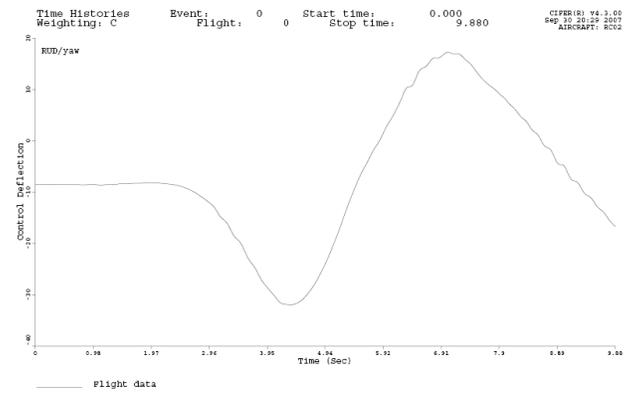

**Figure 13:** Aileron input to verify state space model

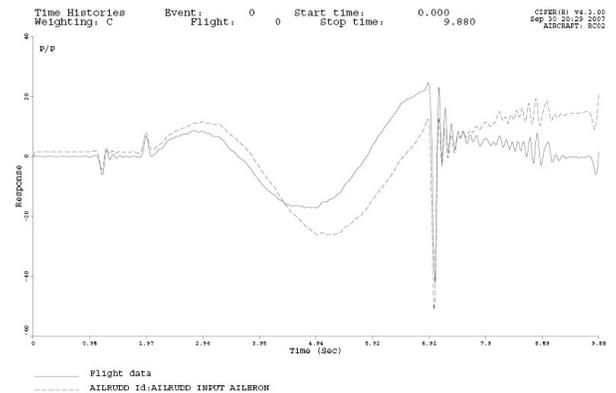

**Figure 14:** Aileron to P from flight data compared with state space model result

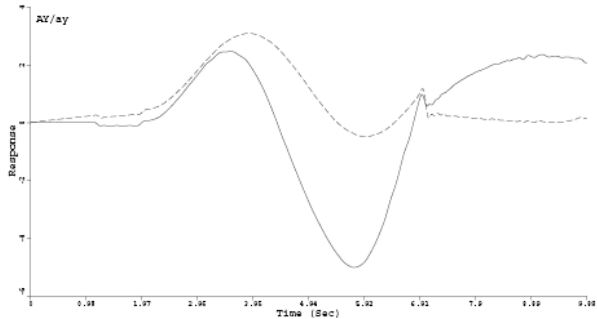

**Figure 15:** Aileron to Ay from flight data compared with state space model result

## 7 Concluding Remarks

There are still many steps to complete this research. But there are several concluding remarks that can be drawn until now, here they are :

1. Simple PID flight controller dan assist the flight data collection for system identification using frequency response analysis. The data (input auto spectrum plots) proved that this system can give very good frequency input coverage which is quite difficult to get using manually controlled flight especially for unmanned system.

2. The data collected give very good coherence for single-input and single-output frequency response.

3. The verification of state space model gave correct qualitative result, but still did not gave satisfactorily result. There are still many experimentation steps to increase the state space model accuracy. Among those alternatives are :

    o Try different window length in FRESPID and COMPOSITE.

    o Increase the frequency of data recording

    o Develop parameterized state space model template using first principle approach

## References

[1] Mark B. Tischler, and Robert K. Remple, *Aircraft and Rotorcraft System Identification : Engineering Methods with Flight Test Examples*, American Institute of Aeronautics and Astronatics, 2006

[2] Bernard Mettler, *Identification Modeling and Characteristics of Miniature Rotorcraft*, Kluwer Academic Publishers, 2003

[3] Bernard Mettler, Takeo Kanade, and Mark B. Tischler, *System Identification Modeling of a Model-Scale Helicopter*, Carnegie Melon University, 2003

[4] Widyawardana Adiprawita, *Development of Simple Autonomous Fixed Wing Unmanned Aerial Vehicle Controller Hardware* (Internal Report), School of Electric Engineering and Informatics, Bandung Institute of Technology, 2006